\documentclass{article}

\PassOptionsToPackage{numbers, compress}{natbib}
\usepackage[final]{neurips_2021}
\usepackage[turnon]{mynotes}
\usepackage{amsthm}
\newtheorem{proposition}{Proposition}

\usepackage[utf8]{inputenc} % allow utf-8 input
\usepackage[T1]{fontenc}    % use 8-bit T1 fonts
\usepackage{hyperref}       % hyperlinks
\usepackage{url}            % simple URL typesetting
\usepackage{booktabs}       % professional-quality tables
\usepackage{amsfonts}       % blackboard math symbols
\usepackage{nicefrac}       % compact symbols for 1/2, etc.
\usepackage{microtype}      % microtypography
\usepackage{xcolor}         % colors

\usepackage{amsmath}

\title{Permutation Equivariant Generative Adversarial Networks for Graphs}

\author{%
  Yoann~Boget \\
 University of Geneva \\  
Geneva School of Business  \\ 
Administration HES-SO \\  
Carouge, Switzerland \\
\href{mailto:yoann.boget@hesge.ch}{yoann.boget@hesge.ch}
\And
Magda Gregorova \\
 Center for Artificial Intelligence \\ 
 and Robotics (CAIRO), FHWS,\\
  Würzburg-Schweinfurt, Germany\\
\href{mailto:magda.gregorova@fhws.de}{magda.gregorova@fhws.de}
\And
Alexandros Kalousis \\
Geneva School of Business \\ 
Administration HES-SO \\  
Carouge, Switzerland \\
\href{mailto:alexandros.kalousis@hesge.ch}{alexandros.kalousis@hesge.ch}

}

\begin{document}

\maketitle

\begin{abstract}
  One of the most discussed issues in graph generative modeling is the ordering of the representation. One solution consists of using equivariant generative functions, which ensure the ordering invariance. After having discussed some properties of such functions, we propose 3G-GAN, a 3-stages model relying on GANs and equivariant functions. The model is still under development. However, we present some encouraging exploratory experiments and discuss the issues still to be addressed.  
\end{abstract}

\section{Introduction}

Molecule generation is one of the most active fields of research developing generative models for graphs. 
One of the most discussed issues in graph generative modeling is the ordering of the representation. 
Nodes and edges in a graph have no prefixed given order. 
However, most models rely explicitly or implicitly on such an ordering. 
Among the permutation invariant generative models, there is no GAN-based model. GANs though present some benefits in comparison with this class of generative model.

We thus propose a GAN-based model for graph generation. Unlike other GAN-based models, our model is invariant to permutations. Different from the other permutation-invariant models, it can evaluate the graph as a whole rather than embeddings of its elements, thanks to the adversarial loss. 

If graphs are objects without a fixed given order. 
Their representation, however, requires an ordering. We indicate the set of all possible permutations by $\Pi = \{ \pi(i) \}_{i=1}^{n!}$ and, when needed, we refer to one specific ordering under permutation $\pi \in \Pi$ of the graph representation using the superscript $\pi$ as in $G^{\pi}$ for instance. 
We define a generative model as invariant if each of generated element in the graph does not depend on the ordering of the other elements. 

First, we briefly review the existing related works (section \ref{sec:review}) within this background. We then present some property of permutation-equivariant  generative functions (\ref{sec:theory}). Then, we present our model (\ref{sec:model}) and finally present some explanatory experiments (\ref{sec:results}). 

\section{Related Works}\label{sec:review}

Most generative models for graphs are auto-regressive, generating one node, edge, and/or block at a time. By definition, auto-regressive models are not permutation invariant. Some use a canonical sequential representation such as the SMILES representation (for instance: \citep{segler2017generating, olivecrona2017molecular, gomez2018automatic,  kusner2017grammar, dai2018syntax}). Others are generating directly some graph objects such nodes or edges \citep{You2018, popova2019molecularrnn, Liao2019,
shi2020graphaf, lim2020scaffold, li2018learning,  khemchandani2020deepgraphmolgen}.

Most existing non-autoregressive models are permutation invarant. Various classes of model have been proposed including  VAE-  \citep{Yang2019, flam2020graph},  Flows- \citep{liu2019graph, madhawa2019graphnvp}, and Score-based \citep{Niu2020} models. They all use permutation equivariant generated functions as detailed in section \ref{sec:equi_func}.

There are few models, which do not fall into these categories. Some VAE-based models let the decoder generate all possible permutations through an MLP \citep{Simonovsky2018, ma2018constrained}. However, they require a matching procedure to compute the reconstruction loss. Finally, two GAN-based models, MolGAN and GG-GAN, use generators that are not invariant to permutations. However, thanks to a permutation invariant discriminator, they do not require a matching procedure as the VAE models do. We discuss further details of the GAN-based model in the section \ref{sec:comparison}. 

\section{Permutation Equivariant Generative Functions}\label{sec:theory}

It is common to define a graph $G = (V, E)$ as a pair made of a set of nodes $V =  \{\nu_1,..., \nu_n \}$, and a set of edges between pairs of the nodes $E = \{(\nu_i, \nu_j) \in V^2\}$. 
In addition, we define the multiset of node attributes $X = \{\{x_1, ... ,x_n \}\}$ and the multiset of the edge attributes between nodes $W = \{\{w_{i, j}, ... ,w_{i, j}: i, j \in \mathbb{N}, i, j \leq n\}\}$ \footnote{Note that we have an attribute even if there is no edge.}. 
We note that a node is fully defined by the couple $g_i = (x_i, \{\{w_{i, 1}, ..., w_{i, n}\}\})$ and, therefore a graph is also fully defined by the multiset $G = \{\{g_i, ..., g_n\}\}$. 
In consequence, we can applied some results shown for (multi)sets to graphs.

A generative function for graph is a function $\mathbf{f}$ outputting sets $G \in \mathcal{G}^n$. We define it as invariant to permutations if $\mathbf{f}(Z^{\pi}) = G^{\pi} \quad \forall \pi \in \Pi$, $Z$ being a set of identically and independently sampled (iid's) variables from a know distribution.

\subsection{Properties of Permutation Equivariant Generative Functions}\label{sec:equi_func}
 
 We built a generative permutation equivariant function as follow. We associate each $z_i \in \mathcal{Z}^n$ with a $g_i \in \mathcal{G}^n$. We define function $f(z_i, Z_{-i}) = g_i$, which is invariant to the ordering in $Z_{-i}$ defined as $Z_{-i}= Z \setminus \{\mathbf{z}_i\}$. We define a function $\mathbf{f}(Z) =\{f(z_i, Z_{-i}): i \in \mathbb{N}, i \leq n\} \}$, which transforms $\mathcal{Z}^n$ to $\mathcal{G}^n$, independently from the ordering in $Z$, and, consequently, from the one in $G$. It is easy to see that $\mathbf{f}(Z)$ is permutation equivariant, i.e. $\mathbf{f}(Z) = G \iff \mathbf{f}(Z^\pi) = G^\pi \forall \pi \in \Pi$. Since it is the same function $f$ applied to all pairs $(z_i, g_i)$, the function $\mathbf{f}$ implies, by construction, parameter sharing. 
 
It is shown \citep{Zaheer2017} and \citep{Wagstaff2019}\footnote{\citet{Zaheer2017} have shown the case for countable elements in $Z$ and \citet{Wagstaff2019} for the continuous case, but with the additional condition that the latent space defined as the domain of $g$ and the codomain of $h$ in equation \eqref{equivariance} is larger or equal to $n$.}, that a function is permutation invariant if and only if it can be decomposed in the form $g \left( \sum_{i} h(x_i) \right)$ for suitable transformations $g$ and $h$ \footnote{The summation can be replaced by some other aggregation functions, such as the maximum or the mean.}. It follows that the function $f$ can be decomposed as:

\begin{equation}\label{equivariance}
    f(\mathbf{z}_i, Z_{-i}) =  g \left(z_i, \sum_{j \neq i} h \left(z_i, z_j; \psi \right) ; \phi \right)
\end{equation}

This decomposition imposes another level of parameter sharing through the function $h$. Hence,  $\mathbf{f}(Z)$ has two levels of parameter sharing, one across the elements in $Z$ through $g$ another within each element through $h$. 

The equiprobability of all the permutations in the ordering of the generated instance is another consequence of equivariance. 

\begin{proposition}
Given a set $Z$ containing some elements $z_i \in \{1, ..., n\}$, which are independent and identically distributed realisations of a random sampling, and a permutation equivariant function $f(Z) = G$ then: $P(G^{\pi(i)}) = P(G^{\pi(j)}) = \frac{1}{n!} \sum_{k = 1}^{n!} p( G^{\pi(k)}) \quad \forall i, j \in \{1, ... ,n!\} $ .
\end{proposition}

\begin{proof}
The proof follows directly from the equiprobabilities of the $Z^\pi$'s. See appendix \ref{proof} for the details.
\end{proof}

To summarize, a permutation equivariant function generating instances from iid $z_i$'s 1) is restricted to the class of functions defined in equation \eqref{equivariance}, 2) implies parameter sharing across and within the elements, 3) enforces equiprobability for each permutation of isomorphic graphs.  

\subsection{Comparison With Other Models}\label{sec:comparison}

For a generic function $\phi: \mathcal{Z} \mapsto \mathcal{G}$, $\mathcal{G}$ being a set of graphs with cardinality $N$.
Then, the codomain of $\phi$ is of cardinality $n! N$ in the general case, but only of cardinality $N$ if $\phi$ is invariant to ordering permutations in $G$. 
In that sense, we can say that the complexity of the mapping between $\mathcal{Z}$ and $\mathcal{G}$ is (much) higher if $\phi$ is not permutation invariant. So, the graph generative models that are not permutation invariant can use a broader class of functions (since they do not need to be of the form of equation \eqref{equivariance}), but their task is of higher complexity.

The two existing generative models based on GANs are not invariant to permutations. Therefore, they lose capacity at learning the multiple permutations of the same graph.

MolGAN uses a Multi-layers Perceptron (MLP) as generator. As a result, the model suffers from frequent mode-collapse, which the authors solve by an early stopping procedure. Despite these efforts, the uniqueness of the generated graphs by their model is still very low ($\approx 3\%$), suggesting that the model collapse is a serious issue of this model. 

GG-GAN uses a permutation equivariant generator. However, the model also takes as input a set of fixed ordered identification vectors, which are concatenated to the $z_i$'s. Consequently, each permutation of the $z_i$'s produces a new set of input and, therefore, the generator loses its equivariance. The authors report the number of isomorphic classes generated, which are not in the training set (out of 5k generated instances). However, it is not clear if they check for validity. If they do, their results are better than the results reported by MolGAN. In the result section, we assume that these classes are valid molecules. 

In comparison with other permutation invariant generative models such as VAE, flows, or score-based,
permutation invariant GANs-based models have an essential benefit. Thanks to the adversarial loss of a permutation-invariant discriminator, they can evaluate the whole graph. On the contrary, invariant VAE, flows, or score-based models evaluate only the log-likelihood of node/edge embeddings (or its gradient) rather than the graph as a whole. 

\section{3G-GAN}\label{sec:model}

We propose 3G-GAN, a permutation invariant GAN-based generative model, made of 3 stages, working as a small auto-regressive model. In addition to the benefits already mentioned, our model relies on a non-sequential decomposition learning distributions assumed to be easier to model. The model is still under development, and we discuss the issues that are still to address in section \ref{sec:issues}. 
However, we already have promising preliminary results presented in the section \ref{sec:results}. 

\subsection{The model}

The first stage model the probability $p_g(A, z)$. The set of binary edge attributes $A$ is similar to $W$ except that the $a_{i, j}$'s contain only the binary information (1 if there is an edge, 0 otherwise) rather than the complete description of the edge attributes as in the $w_{i,j}$'s. The set $A$ represents the skeleton of a molecule. The second stage is modeling the nodes attributes given the skeleton $p(X|A, z)$ and the third stage is the edge attributes given the skeleton, and the node attributes $p(W|X, A, z)$. We give the complete objective function in appendix \ref{objective}. 

In our current implementation, we engage the three stages independently during training using the real data (rather than the previously generated data) for conditioning, following teacher-forcing strategy. 
At generation time, we follow sequentially the 3 stages, generating first the set of binary edges $A$, generating then the set of nodes $X$ conditioning on these matrices, and finally generating the edge attribute $W$ conditioning on the previous two steps.

The permutation equivariance is implemented through GNNs, similar to InteractionNet (\citet{Battaglia2016}). Details about the architectures are given in appendix \ref{archi}. 

In practice for molecules, we encode the information of $A$, $X$, and $W$ into tensors. The set $A$ is encoded as an unweighted and undirected adjacency matrix. An annotation matrix encodes the atom types $X$, each row of the matrix being a one-hot vector,  the bond types (single, double, triple, aromatic) in $W$ are encoded into a tensor, each $w_{i,j}$ being a one-hot vector if there is a bond, and a vector of $\mathbf{0}$ otherwise. 

\subsection{Issues}\label{sec:issues}
So far, we have developed the last two stages. Using the skeleton $A$ from the data, we can generate the atom and bond types and so create new molecules. In the next section, we report the results of this experiment.

Modeling the skeleton $p_g(A, z)$ requires to address some additional issues. We pointed out two that are major issues for permutation invariant generative models and should be a concern for the community. 

The first issue concerns the GNNs having as input a graph without neither node nor edge attributes, as our discriminator in the first stage. We know from \citep{Morris2019, Loukas2019} that without attributes, which can serve as node or edge identification, some non-isomorphic graphs are indistinguishable by most classical graph neural networks, including Message Passing Neural Networks (MPNN) \citep{Gilmer2017}. However, the effect of these limitations have not been studied for generative functions. 

Second, generating $A$ requires to use a function without information about the graph topology, which is restricting even further the class of function $f$ described in equation \eqref{equivariance}\footnote{Typically, MPNN uses the topological information as masking, setting $h$ to $0$, so that the aggregation function is over the neighborhood of node $i$ rather than over all nodes in the graph.}. 
Only a few works about graph generative modeling  \citep{Yang2019, flam2020graph, Niu2020} have addressed this specific issue.

\section{Experiments} \label{sec:results}

We report here an exploratory experiment. We train stages 2 and 3 as described in the previous section. For generation, we need a sample of sets $A$. Instead of generating it through a parametric model, we sample the sets $A$ from from the data.

We test our model on the QM9 dataset \citep{ramakrishnan2014quantum}. We stress that, unlike most, if not all existing models for molecule generation, our model generate the number of hydrogen atoms linked to a heavy one, the formal charge\footnote{To do so, we encode the formal charge and the number of hydrogen into the node attribute, so that we have 21 atom types instead of usually 4} as well as the aromatic bonds. So, we do not rely on chemoinformatics software to fix these properties. It can explain the relatively low validity rate in our experiment. We use RDKit only for preprocessing and final evaluation metric calculation. Therefore, the comparison with other models is not fair \footnote{In addition, we note that both MolGAN and GG-GAN help their discriminators by adding various handcrafted features. On our side, the use of skeleton directly from the data is a significant 
advantage.}.

We use the classical metrics for \textit{de novo} molecule generation: validity (val), uniqueness  (uniq) and novelty (nov). 
We add to these the global rate of valid-unique-novel molecules (\emph{all}). 
We report novelty even if it is not the most relevant metric for evaluating the QM9 dataset since it is supposed to contain most organic molecules.

We report the main results comparing the generative performance of our method to other GAN-based graph generating models in Table \ref{tab:1}. Our model outperforms both MolGAN and GG-GAN in the global rate of valid-unique-novel generated molecules. Though our validity rate is worse than the one of MolGAN, which optimizes the validity rate implicitly through a Reinforcement Learning objective, this relative weakness of our model is more than compensated by the much higher diversity of the generated molecules. We cannot provide a similar discussion for the GG-GAN comparison as the paper does not release the values for the individual metrics.

\begin{table}[ht]
\caption{Comparison Between GAN-based Models}\label{tab:1}
\begin{center}
\begin{tabular}{lllll }

\textbf{Model} &  \textbf{val} & \textbf{uniq} & \textbf{nov} & \textbf{all}   \\
\hline \\
MolGAN RL & 99.8 & 2.3 & 97.9 & 2.2 \\
MolGAN no RL  & 87.7 & 2.9 & 97.7 & 2.5 \\
GG-GAN \footnote{The code of GG-GAN is not public yet and they only report the number of 'isomorphism classes'. We assume that these classes are valid molecules even if it was not specified so.} & & & & 16.6 \\
3G-GAN (ours) & 78.5 & 53.9 & 62.0 & 26.2 \\

\end{tabular}
\end{center}
\end{table}

\section{Conclusion}

We proposed a model called 3G-GAN, a non-sequential auto-regressive GAN-based model combining the advantages of permutation-invariant generative models and those of the adversarial training. Still under development, we provide encouraging results from exploratory experiments.

\bibliographystyle{apalike}
\bibliography{references}
\clearpage
\appendix

\section{Proofs of proposition 1}\label{proof}

\paragraph{Proposition:}
Given a set $Z$ containing some elements $z_i \in \{1, ..., n\}$, which are independent and identically distributed realisations of a random sampling, and a permutation equivariant function $f(Z) = G$ then: $P(G^{\pi(i)}) = P(G^{\pi(j)}) = \frac{1}{n!} \sum_{k = 1}^{n!} p( G^{\pi(k)}) \quad \forall i, j \in \{1, ... ,n!\} $ .

\paragraph{Proof:}
\begin{equation*}
\begin{split}
P( G^{\pi(i)}) &= P(f^{-1}(G^{\pi(i)})) \\
& = P(Z^{\pi(i)} \in  f^{-1}(G^{\pi(i)})) \\
& = P(Z^{\pi(j)} \in  f^{-1}(G^{\pi(j)}))  \\
& = P( G^{\pi(j)}) \\
\end{split}
\end{equation*}

The third equality holds because $P(Z^{\pi(i)}) = P(Z^{\pi(j)})$ and $f$ is permutation equivariant. Note that the preimage $f^{-1}(G)$ is well defined even if $f$ is not invertible. 

From this result, we have:

\begin{equation*}
\begin{split}
\sum_{k = 1}^{n!} P(G^{\pi(k)}) 
 = n! P( G^{\pi(i)}) \iff P( G^{\pi(i)}) = \frac{1}{n!} \sum_{k = 1}^{n!} p( G^{\pi(k)}) \quad \forall i \in \{1, ... ,n!\} 
\end{split}
\end{equation*}

\section{Objective function of the 3G-GAN}\label{objective}

We give here the 3 objective functions for the 3 stages of our model. The functions $d$ and $g$ are respectively the discriminator and the generator. 

\paragraph{Stage 1}
\begin{equation*}
\min\limits_g\max\limits_d V(d, g)=  \mathbb{E}_{A \sim \mathbf{p}_{data}} 
\left[- d \left( A \right) \right] + \mathbb{E}_{\mathbf{z}\sim \mathbf{p}_{\mathbf{z}}}
\left[ d \left( g \left( \mathbf{z} \right) \right) \right]
\end{equation*}

\paragraph{Stage 2}
\begin{equation*}
\min\limits_g\max\limits_d V(d, g)=  \mathbb{E}_{(X, A) \sim \mathbf{p}_{data}} 
\left[  - d \left( X, A \right) \right] + \mathbb{E}_{\mathbf{z}\sim \mathbf{p}_{\mathbf{z}}, A \sim \mathbf{p}_{data}}
\left[ d \left( g \left( \mathbf{z}, A \right), A \right) \right]
\end{equation*}

\paragraph{Stage 3}
\begin{equation*}
\min\limits_g\max\limits_d V(d, g)=  \mathbb{E}_{(X, W) \sim \mathbf{p}_{data}} 
\left[  - d \left( X, W \right) \right] + \mathbb{E}_{\mathbf{z}\sim \mathbf{p}_{\mathbf{z}}, (A, W) \sim \mathbf{p}_{data}}
\left[ d \left( g \left( \mathbf{z}, X, W \right), X, W \right) \right]
\end{equation*}

\section{Architecture of our Permutation-Equivariant GNN}\label{archi}

 Using $\mathbf{h}_{i}^{(l)}$ and $\mathbf{r}_{i, j}^{(l)}$ as the representation of respectively the node $i$ and the edge between nodes $i,j$ in the $l$-th layer, the update for both embeddings is as follow:

\begin{equation}\label{eq:9}
\mathbf{r}_{i, j}^{'(l+1)} = f_{r'}([\mathbf{h}_{i}^{(l)}, \mathbf{h}_{j}^{(l)}])
\end{equation}

\begin{equation}\label{eq:10}
\mathbf{r}_{i, j}^{(l+1)} =  f_{r}([\mathbf{r}_{i, j}^{(l)}, \mathbf{r}_{i, j}^{'(l+1)}])
\end{equation}

\begin{equation}\label{eq:11}
\mathbf{h}_{i}^{(l+1)} = \sigma(f_h([\mathbf{h}_i^{(l)}, \sum_{j \in \mathcal{N}(i)}  \mathbf{r}_{i, j}^{(l+1)}]))
\end{equation}

We use linear transformations as update functions $f$ and add a CELU non-linearity (noted $\sigma$) after the node update function $f_h$. 

\end{document}